\documentclass{Interspeech}

\usepackage{graphicx}
\usepackage{booktabs}
\usepackage{amsmath}
\usepackage[T1]{fontenc}
\usepackage[utf8]{inputenc}
\usepackage{url}

\begin{document}
\nolinenumbers   

\title{Pashto Common Voice: Building the First Open Speech Corpus\\
       for a 60-Million-Speaker Low-Resource Language}

\name{Hanif Rahman, Shafeeq ur Rehman}
\address{Pashto DAO}

\maketitle

\begin{abstract}
We present the Pashto Common Voice corpus --- the first large-scale,
openly licensed speech resource for Pashto, a language with over 60 million
native speakers largely absent from open speech technology. Through a
community effort spanning 2022--2025, the corpus grew from 1.5~hours and
5~contributors to 147~total hours and 1,483~unique speakers across ten
Mozilla Common Voice releases (CV14--CV23). Speaker participation increased
approximately 108-fold between CV17 and CV18, coinciding with a VOA Pashto broadcast
campaign. We describe the full methodology: interface localisation,
Wikipedia-based sentence extraction with automated filtering, phonemically
targeted contributions for the four most frequently dropped Pashto
characters, and multi-channel community outreach. MCV23 contains 107,781~clips (60,337
validated; 82.33~validated hours) across 13~content domains. Fine-tuning
Whisper Base on the MCV20 yields \textbf{13.4\% WER} on the
MCV20 test split, against the published Whisper Base zero-shot WER of 99.0\%
on Pashto.
\end{abstract}


\section{Introduction}

Pashto is spoken by an estimated 60--80 million people across Afghanistan,
Pakistan, and the diaspora. Despite this scale, the language has been largely
absent from open speech technology: prior to this work, no freely licensed
Pashto speech corpus existed at sufficient scale to train or fine-tune
competitive ASR systems. Small proprietary datasets have appeared over two
decades~\cite{bisani2004speech}, but none were available to the wider
research community. The language's phonological complexity compounds this: eight consonants
absent from Arabic and Persian keyboards are routinely dropped or substituted
in informal Pashto text, so both training data and pre-trained models
transferred from related languages fail on these sounds.

This paper documents the creation and growth of the Pashto corpus on
Mozilla Common Voice (MCV), from initial platform localisation in 2022
through the MCV23 release in September 2025. Our contributions are:
\begin{enumerate}
  \item The first complete account of a sustained, multi-phase effort to
        build an open Pashto speech corpus: from zero to 147~hours and
        1,483~contributors across ten MCV releases.
  \item A quantified account of community growth dynamics, including an
        \textbf{$\sim$108-fold increase in speaker participation} between
        consecutive releases tied to a broadcast media campaign.
  \item A methodology for phonemically targeted sentence curation addressing
        the Pashto consonants most frequently dropped in informal digital text,
        applicable to other script-minority languages with similar keyboard gaps.
  \item A full characterisation of MCV23 Pashto across scale, quality,
        domain coverage, and speaker demographics.
  \item A first-party ASR baseline: Whisper Base fine-tuned on MCV20
        achieves \textbf{13.4\% WER}~\cite{rahman2025psbasel1} on the MCV20
        test split, establishing a reference point for future Pashto ASR work
        on this corpus.
\end{enumerate}

\section{Background and Related Work}

\subsection{Pashto: Linguistic Context}

Pashto is an Eastern Iranian language written in a 45-letter Perso-Arabic
abjad, with two major dialect groups: Southern (Kandahari) and Northern
(Yusufzai), each with divergent pronunciation norms and competing orthographic
standards. Pashto has eight consonants absent from both Arabic and Persian:
four retroflex stops and nasals (\textit{dal}, \textit{te}, \textit{nun},
\textit{re}) and four fricatives and affricates (\textit{shin}, \textit{zhe},
\textit{dze}, \textit{tse}). Because standard Arabic and Persian keyboards
omit all eight, informal digital Pashto text frequently substitutes the
nearest Arabic lookalike, producing training prompts mismatched to actual
spoken Pashto. The fricatives and affricates are especially susceptible:
they have no visually similar Arabic counterpart and are most often dropped
entirely.

\subsection{Prior Work}

Pashto NLP and ASR research has been limited by the absence of public data;
a recent systematic survey~\cite{khan2024survey} traces this gap across
corpora, tools, and models. Bisani et al.~\cite{bisani2004speech} worked
with a small proprietary corpus for speech translation as far back as 2004. The NLPashto
toolkit~\cite{jawad2023nlpashto} covers morphological analysis but not speech.
Transformer-based ASR models have been reported for
Pashto~\cite{zaidi2024pashto}, though data constraints kept training sets
small. Shah et al.~\cite{shah2025benchmarking} evaluated Whisper across
Pashto, Punjabi, and Urdu in an N-shot setting, reporting a Whisper Base
zero-shot WER of 47.3\% on naturally occurring Pashto speech, improving to
41.2\% with 10-shot fine-tuning. The present corpus is the first open resource
at a scale that enables full fine-tuning rather than few-shot adaptation.

\subsection{Mozilla Common Voice}

Mozilla Common Voice (MCV) is an open crowdsourced speech platform releasing
datasets under the CC-0 license~\cite{ardila2020common}. Crowdsourced
initiatives of this kind have become the primary route to new speech data
for dozens of low-resource languages~\cite{vanesch2020google}. Speakers
read sentences aloud; other community members validate recordings through a
dual-review gate (two independent upvotes to validate, two downvotes to
reject). Datasets are released quarterly, each versioned. For low-resource
languages, MCV offers a practical path to freely redistributable speech data
with no licensing overhead, which matters for diaspora communities where
institutional support is absent.

\section{Data Collection Methodology}

\begin{figure}[t]
  \centering
  \includegraphics[width=\columnwidth]{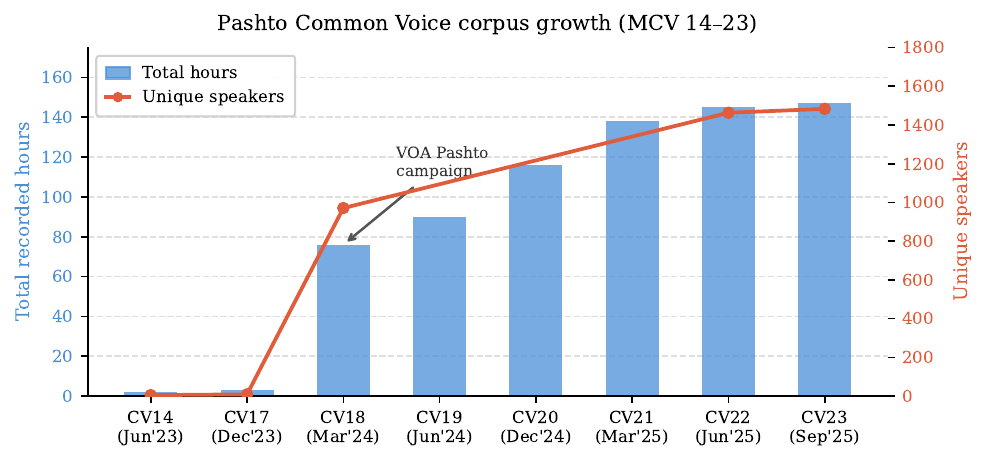}
  \caption{Pashto Common Voice corpus growth across MCV releases 14--23,
           showing total hours (bars) and unique speakers (line). The
           CV17$\!\to$CV18 inflection (Mar.--Jun.~2024) is attributable
           to the VOA Pashto media campaign.}
  \label{fig:growth}
\end{figure}

Corpus construction proceeded in seven phases over three years.
Figure~\ref{fig:growth} shows the resulting growth trajectory.

\textbf{Phase~1 --- Interface Localisation (2022--Feb.~2023).}
MCV had no Pashto support. The organizing team translated the full web interface (over 1,200 strings)
through Mozilla's Pontoon localisation platform. The Pashto locale went live in February 2023. Without this step,
Pashto speakers had no readable interface to contribute through.

\textbf{Phase~2 --- Initial Sentence Corpus (Feb.~2023).}
An initial set of 1,053~sentences was curated via Mozilla's Sentence
Collector from public-domain texts, selected for phonetic diversity and
common Pashto vocabulary.

\textbf{Phase~3 --- Wikipedia Sentence Extraction (2023).}
The organizing team developed a pipeline to extract approximately 27,000
sentences from Pashto Wikipedia (CC-licensed). Each sentence was
automatically filtered to: (a)~remove special characters, digits, and
foreign-language tokens; (b)~retain only sentences of 15~words or fewer,
ensuring suitability for read-speech recording. Extracted sentences were ingested manually in batches of $\sim$100 via the
MCV web interface, allowing a quality-control pass before each batch. This
traded throughput for text fidelity. The pipeline ran continuously alongside
other methods throughout the collection period. The batch included Pashto proverbs scraped from Wikipedia,
prioritised for their short length, lexical diversity, cultural salience, and
free licensing. Of the $\sim$27,000 extracted sentences, approximately 5,000 passed the MCV
community sentence-review stage and are reflected in the corpus; the remainder
were rejected during review.

\textbf{Phase~4 --- Phonemically Targeted Contributions (2023--2024).}
Analysis of the accumulated corpus revealed sparse coverage of the four
fricative and affricate characters (\textit{shin, zhe, dze, tse}), which are
most frequently dropped in informal digital Pashto. The four retroflex stops
and nasals (\textit{dal, te, nun, re}) were present but also under-represented
relative to their frequency in spoken Pashto. Ahmad Shah Azmi provided
sentences specifically covering each of the four fricatives and affricates
across multiple phonetic contexts, giving the corpus reliable training signal
for the sounds that most sharply distinguish Pashto from Arabic and Persian.

\textbf{Phase~5 --- Sentence Expansion (2023--2024).}
The sentence pool grew continuously through further Wikipedia extraction and
community contributions. Ahmad Wali Achakzai of Qamosona.com independently
contributed thousands of sentences through the MCV web interface, bringing
vocabulary coverage from a resource developed specifically for Pashto
lexicography. By MCV23, the validated sentence pool reached 25,109~entries
across 13~content domains (Table~\ref{tab:stats}).

\textbf{Phase~6 --- Community Outreach and Media Campaign (2023--2025).}
The Pashto DAO Facebook page ran as the project's public-facing channel
throughout the collection period, with posts tied to each quarterly release:
announcing milestones, recruiting volunteers, and linking to the recording
interface. The campaign was framed around accessibility rather than data collection:
posts explained that the corpus would power open-source Pashto voice
technology for people who are blind, disabled, non-literate, or do not read
English. The reasoning was that a concrete, community-relevant benefit would
resonate more with diaspora audiences than an abstract appeal to contribute
to AI training data, though no controlled comparison was run. The first public call went out on
20~February 2023, the day the locale launched. The official recording launch
was announced on 12~March~2023.

To reduce friction for non-technical volunteers, the team produced tutorial
videos in Pashto covering two tasks: adding sentences to the collector, and
recording and validating audio. These were distributed through the Facebook
page and substantially lowered the barrier to first contribution.

The largest single growth event came from broadcast media. A community
co-organizer arranged interviews with two VOA Pashto journalists: Qasim Khan
Mandokhel, who produced the initial video segment reaching a global Pashto
audience, and Abdul Hai Kakar, whose follow-up coverage extended the momentum.
The result is visible in the corpus data: between CV17 (March~2024, 9~speakers)
and CV18 (June~2024, 971~speakers) speaker count increased approximately 108-fold
(Figure~\ref{fig:growth}). No platform changes occurred during this period;
the growth coincides with the VOA coverage. That broadcast media via an
engaged community intermediary can produce order-of-magnitude volunteer growth
is the paper's primary transferable finding.

\textbf{Phase~7 --- Progressive Releases (CV14--CV23).}
Each quarterly MCV release incorporated new recordings and re-validated
existing clips. Table~\ref{tab:releases} shows the complete release history
across all ten releases.

\begin{table}[t]
  \centering
  \caption{Pashto MCV corpus release history.}
  \label{tab:releases}
  \begin{tabular}{@{}llrr@{}}
    \toprule
    Release & Date & Hours & Speakers \\
    \midrule
    CV14 & Jun.~2023 &   1.49 &      5 \\
    CV15 & Sep.~2023 &   1.66 &      7 \\
    CV16 & Dec.~2023 &   1.71 &      8 \\
    CV17 & Mar.~2024 &   2.11 &      9 \\
    CV18 & Jun.~2024 &  75.94 &    971 \\
    CV19 & Sep.~2024 &  89.72 & 1{,}098 \\
    CV20 & Dec.~2024 & 115.13 & 1{,}231 \\
    CV21 & Mar.~2025 & 137.33 & 1{,}435 \\
    CV22 & Jun.~2025 & 144.73 & 1{,}463 \\
    CV23 & Sep.~2025 & 147.07 & 1{,}483 \\
    \bottomrule
  \end{tabular}
\end{table}

\section{Corpus Characterisation}

\subsection{MCV23 Core Statistics}

Table~\ref{tab:stats} summarises the MCV23 Pashto subset, released
September~2025. The Dev/Test/Train split sizes deserve a note: MCV constructs
these from a speaker-disjoint subset of validated clips; the much larger
validated pool (60,337~clips) is available as additional training data. The train split being smaller than dev and test in clip count is an artefact
of MCV's speaker-balancing algorithm~\cite{ardila2020common}, not a
labelling error.

\begin{table}[t]
  \centering
  \caption{MCV23 Pashto corpus statistics.}
  \label{tab:stats}
  \begin{tabular}{@{}lr@{}}
    \toprule
    Statistic & Value \\
    \midrule
    Total clips         & 107{,}781 \\
    Validated clips     &  60{,}337 \\
    Validated hours     &   82.33~h \\
    Total hours         &  147.07~h \\
    Unique speakers     &   1{,}483 \\
    Validated sentences &  25{,}109 \\
    Content domains     &       13 \\
    Archive size        &   2.73~GB \\
    Avg.\ clip duration &    4.91~s \\
    Dev / Test / Train  & 3{,}660 / 3{,}660 / 4{,}693 \\
    License             & CC-0 \\
    \bottomrule
  \end{tabular}
\end{table}

\subsection{Growth Dynamics}

Figure~\ref{fig:growth} charts growth across all ten releases.
CV14--CV17 (June~2023 to March~2024) produced only 2.1~hours total
and never exceeded 9~speakers. The CV17$\to$CV18 transition is a step
discontinuity: 75.9~hours and 971~speakers in a single cycle, coinciding
with the VOA broadcast campaign. From CV18 onward, growth was steady but slower: the corpus added 71~hours
and roughly 500~speakers across five releases, with speaker count plateauing
near 1,463--1,483 by CV22--CV23. The initial recruitment cohort had saturated;
sustaining further growth will require a new outreach push.

One structural opportunity remains: 42,265~clips (39.2\% of the total)
are unreviewed as of MCV23, representing an estimated 57~additional
validated hours if the reviewer pool can be mobilised.

\subsection{Domain and Demographic Distribution}

\begin{figure}[t]
  \centering
  \includegraphics[width=\columnwidth]{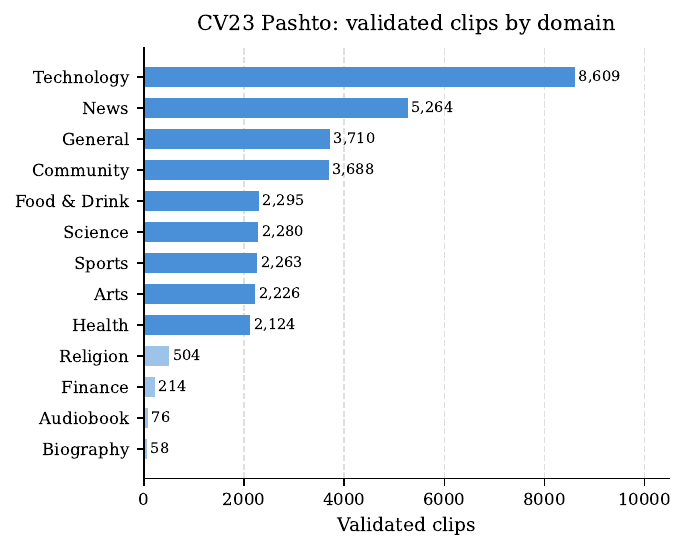}
  \caption{MCV23 Pashto: validated clips by content domain.}
  \label{fig:domains}
\end{figure}

Figure~\ref{fig:domains} shows the distribution of validated clips across
13~content domains. Technology (8,609~clips) and News (5,264) dominate,
reflecting the platform's default sentence corpus. Religion (504), Finance
(214), Audiobook (76), and Biography (58) are under-represented and are
targets for future sentence collection.

\begin{figure}[t]
  \centering
  \includegraphics[width=0.85\columnwidth]{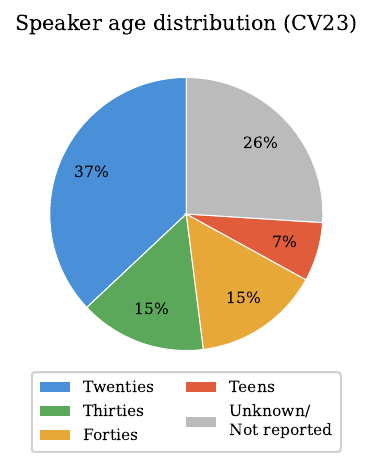}
  \caption{MCV23 speaker age distribution. Gender is unreported for
           98\% of speakers (a critical metadata gap).}
  \label{fig:demographics}
\end{figure}

Figure~\ref{fig:demographics} shows the age distribution. The contributor
base skews young: 37\% in their twenties, 15\% thirties, 15\% forties,
7\% teenagers. The more pressing gap is gender: 98\% of speakers did not report gender,
making any assessment of gender balance impossible. The corpus currently
cannot support gender-aware evaluation or bias auditing, which limits its
utility for applied ASR work.

\section{Quality Control}

MCV's dual-review quality gate requires two independent upvotes to validate
and two downvotes to invalidate a recording. As of MCV23:
\begin{itemize}
  \item \textbf{Validated:} 60,337 clips (56.0\%)
  \item \textbf{Invalidated:} 5,179 clips (4.8\%)
  \item \textbf{Unreviewed:} 42,265 clips (39.2\%)
\end{itemize}
The 56\% validation rate compares favourably with early-stage MCV datasets
for other low-resource languages~\cite{ardila2020common}. The high unreviewed fraction reflects a volume bottleneck: recordings have
accumulated faster than the reviewer pool can process them, not a quality
failure.

Pashto's orthographic variation poses a sentence-level quality risk:
Pashto's eight keyboard-absent characters are routinely substituted or
dropped in informal text, and prompts containing substitutions will elicit
mispronounced recordings. The dual community review of text prompts before
they enter the recording queue mitigates this, as does the phonemically
targeted sentence contribution strategy described in Phase~4.

\section{Downstream Impact}

The corpus has already enabled ASR work that was not possible before. The original Whisper paper~\cite{radford2022robust} reports 99.0\% WER for
Whisper Base on Pashto zero-shot (Fleurs benchmark), effectively chance-level
performance confirming that the pre-trained model carries no useful Pashto
signal. Shah et al.~\cite{shah2025benchmarking}
report 47.3\% WER zero-shot on naturally occurring Pashto speech (CHiPSAL
benchmark), improving to 41.2\% with 10-shot fine-tuning. MCV20 is available via the Mozilla Common Voice
platform~\cite{ardila2020common} at \texttt{commonvoice.mozilla.org/ps}.

To establish a full fine-tuning baseline on the MCV data itself, the authors
trained Whisper Base (72.6M parameters) on the MCV20 train split
(4,693~clips). Training ran for 4,900~steps ($\approx$3~epochs) with linear warmup to a
peak learning rate of $5\!\times\!10^{-5}$ at step~1,000, then linear decay,
on a consumer Apple MacBook Pro. Whisper's multilingual BPE tokenizer covers
the Arabic Unicode block; all four Pashto-specific characters are encodable
in the output vocabulary, as confirmed by the model's ability to produce
them correctly during evaluation. Full training configuration is documented
in the model repository~\cite{rahman2025psbasel1}. The
model (\texttt{ihanif/ps\_base\_l1}~\cite{rahman2025psbasel1}) reaches
\textbf{13.4\% WER} on the MCV20 test split (14.6\% orthographic WER).
Table~\ref{tab:wer} shows the WER trajectory.

\begin{table}[t]
  \centering
  \caption{WER (\%) of \texttt{ps\_base\_l1} on the MCV20 Pashto test
           split. Step~100 is the first logged checkpoint (not zero-shot).
           The Whisper Base zero-shot WER on Pashto is 99.0\% on
           Fleurs~\cite{radford2022robust}; our fine-tuned model is
           evaluated on the MCV20 test split.}
  \label{tab:wer}
  \begin{tabular}{@{}rrrr@{}}
    \toprule
    Step & Epoch & WER & WER\textsubscript{ortho} \\
    \midrule
    100  & 0.1 & 93.5 & 94.6 \\
    500  & 0.3 & 57.0 & 60.4 \\
    1000 & 0.6 & 46.7 & 50.6 \\
    2000 & 1.3 & 30.9 & 34.5 \\
    3000 & 1.9 & 22.5 & 25.3 \\
    4000 & 2.6 & 16.5 & 18.3 \\
    \textbf{4900} & \textbf{3.1} & \textbf{13.4} & \textbf{14.6} \\
    \bottomrule
  \end{tabular}
\end{table}

A direct comparison across these figures requires care: Radford et al.\
evaluate on Fleurs, Shah et al.\ on naturally occurring CHiPSAL speech, and
our model on MCV read speech. The 99.0\%\,$\to$\,13.4\% improvement reflects
what full fine-tuning on MCV20 contributes relative to a completely untrained
baseline; the gap versus Shah et al.'s 41.2\% reflects both additional
training data and an easier test condition (read vs.\ naturally occurring
speech). With full access to the MCV20 training split, Whisper Base can be adapted
to practical Pashto ASR on consumer hardware. The MCV23 corpus, 27\%
larger than MCV20, should lower this figure further. The corpus also
provides sufficient audio for self-supervised pre-training frameworks such
as wav2vec~2.0~\cite{baevski2020wav2vec}, which we leave as a direct
comparison for future work.

\section{Discussion}

\textbf{Replicability.}
The collection methodology relies on no proprietary tools beyond the MCV
platform itself. The sentence extraction pipeline (Wikipedia scrape, character
filtering, length capping, manual batch QC) is straightforward to re-implement.
The key lesson is about sequencing: build a digital community presence first
(a social media page in the target language, accessibility-framed messaging,
tutorial videos), then find someone with the right connections to arrange
broadcast coverage. The data suggest the broadcast intervention had far more
leverage than any technical step, but it worked because a community was
already there to receive it.

\textbf{The language technology broker.}
The co-organizer's role (identifying journalists, framing the story for each
outlet, managing relationships) does not fit neatly into standard crowdsourcing
frameworks, which tend to model contribution as an individual act. It is closer
to what the knowledge mobilisation literature calls a \textit{knowledge
broker}~\cite{ward2009knowledge}: an intermediary who connects technical work
to a community that would not otherwise encounter it. For low-resource language
projects, this role can be as consequential as the technical infrastructure.
The person filling it needs credibility in the community, media contacts, and
enough understanding of the project to explain it accessibly. These skills are
rare, and projects should look for them deliberately rather than hoping a
co-author happens to have them.

\textbf{Limitations.}
At 147~hours, the corpus is modest next to high-resource benchmarks (English
exceeds 2,700~hours on MCV). All recordings are prompted read speech, so the
corpus is not directly suitable for conversational ASR. Dialectal coverage is
uncontrolled: contributors are drawn primarily from the diaspora, likely
over-representing Northern (Yusufzai) varieties. Gender metadata is almost
entirely absent, which prevents gender-aware evaluation and bias auditing.
Each of these should be an explicit target for future collection rounds.

\section{Conclusion}

We have documented the creation and growth of the Pashto Common Voice
corpus, the first open speech resource for a language spoken by over
60~million people. On the technical side, a corpus at this scale makes
Whisper Base fine-tuning to 13.4\% WER achievable on consumer hardware,
a result that simply was not possible before. On the practical side, the
dominant growth driver was not a software improvement but a single broadcast
media campaign, arranged by a community intermediary who knew which journalists
to call. For teams building corpora for similarly under-resourced languages,
finding that person early may matter more than optimising the pipeline.
The MCV23 corpus is openly available at \texttt{commonvoice.mozilla.org/ps}
under CC-0 licence.

\section{Acknowledgements}

We thank Ahmad Shah Azmi for targeted phonemic sentence contributions
covering the four fricative and affricate characters; Ahmad Wali Achakzai
(Qamosona.com) for community awareness contributions; Qasim Khan Mandokhel
and Abdul Hai Kakar of VOA Pashto for their journalism, which produced the
single largest growth event in the corpus; Noor Ullah Atal, Pashto linguist,
for guidance throughout the project; and the Mozilla Foundation for the
Common Voice platform.

\textit{AI tool disclosure: Portions of this manuscript were drafted
with the assistance of Claude (Anthropic).}


\end{document}